\documentclass[10pt,twocolumn,letterpaper]{article}

\usepackage{iccv}
\usepackage{times}
\usepackage{epsfig}
\usepackage{graphicx}
\usepackage{amsmath}
\usepackage{amssymb}

\usepackage[pagebackref=true,breaklinks=true,letterpaper=true,colorlinks,bookmarks=false]{hyperref}

\iccvfinalcopy 

\def\iccvPaperID{29} 

\ificcvfinal\fi

\begin{document}

\title{Memory Population in Continual Learning via Outlier Elimination}

\author{Julio Hurtado\textsuperscript{1}, Alain Raymond-Sáez\textsuperscript{2}, Vladimir Araujo\textsuperscript{2,3},\\ Vincenzo Lomonaco\textsuperscript{1}, Alvaro Soto\textsuperscript{2}, Davide Bacciu\textsuperscript{1}\\
\textsuperscript{1}University of Pisa, \textsuperscript{2}Pontificia Universidad Católica de Chile, \textsuperscript{3}KU Leuven \\
{\tt\small julio.hurtado@di.unipi.it, afraymon@uc.cl, vgaraujo@uc.cl,} \\{\tt\small vincenzo.lomonaco@unipi.it, asoto@ing.puc.cl, davide.bacciu@unipi.it}
}

\maketitle
\ificcvfinal\thispagestyle{empty}\fi

\begin{abstract}
Catastrophic forgetting, the phenomenon of forgetting previously learned tasks when learning a new one, is a major hurdle in developing continual learning algorithms. 
A popular method to alleviate forgetting is to use a memory buffer, which stores a subset of previously learned task examples for use during training on new tasks.
The \textit{de facto} method of filling memory is by randomly selecting previous examples. However, this process could introduce outliers or noisy samples that could hurt the generalization of the model.
This paper introduces Memory Outlier Elimination (MOE), a method for identifying and eliminating outliers in the memory buffer by choosing samples from label-homogeneous subpopulations. 
We show that a space with a high homogeneity is related to a feature space that is more representative of the class distribution. In practice, MOE removes a sample if it is surrounded by samples from different labels.
We demonstrate the effectiveness of MOE on CIFAR-10, CIFAR-100, and CORe50, outperforming previous well-known memory population methods.
\end{abstract}

\section{Introduction}

Deep learning models have demonstrated impressive performance in a range of tasks, including image recognition \cite{he2016deep, dosovitskiy2020image}, natural language processing \cite{kenton2019bert, brown2020language}, and even complex games like Go \cite{silver2016mastering}, and Starcraft II \cite{vinyals2019grandmaster}. However, a major limitation of these models is their lack of versatility - when trained to perform on a sequence of tasks, they often forget how to solve previously learned tasks, a phenomenon known as \textit{catastrophic forgetting}.

Continual Learning (CL)  \cite{parisi2019continual, delange2021continual} methods aim to address Catastrophic Forgetting (CF) by enabling deep learning models to learn new tasks without forgetting previously learned ones. Most of CL approaches can be divided into multiples categories, these include subdividing model parameters into subspaces for each new task \cite{rusu2016progressive, mallya2018piggyback}, imposing constraints on the learned gradients \cite{kirkpatrick2017overcoming, lopez2017gradient}, and using meta-learning to learn reusable weights for all tasks \cite{rajasegaran2020itaml, hurtado2021optimizing}. Out of these categories, memory-based methods such as Experience Replay (ER) \cite{rolnick2019experience, 10.1145/3147.3165} provide a straightforward solution that achieves good results. These methods use a memory to store previous task samples to prevent forgetting during training of the current task. 

However, understanding which elements to put in this memory is still an open question. Recent studies \cite{chaudhry2018riemannian, wu2019large, hayes2020remind, araujo-etal-2022-mem, peng2023ideal} on memory population have found that randomly selecting elements to populate the memory performs nearly as well as more complex selection methods, without requiring additional computation. However, random selection may potentially include noise or outliers that are not useful for generalization, particularly when there are constraints on the memory size. Thus choosing the right elements to train on becomes critical.

To address the aforementioned problems, we propose to eliminate outliers from memory and choose more representative samples from the distribution when facing restriction on the amount of data we can store. Based on the idea that the data distribution is a mixture of subpopulations \cite{Feldman20}, our approach prioritizes samples that are surrounded mainly by samples with the same class label in an embedding space. Outliers and noisy samples will either be far from other samples or surrounded by samples from different classes \cite{Arplt2017}. Thus, selecting samples from label-homogeneous neighborhoods should reduce the appearance of outliers, fomenting the selection of more representative and easier samples to learn by models.

Our method, Memory Outlier Elimination (MOE), applies this criterion before randomly selecting from the remaining samples. We test MOE on standard CL benchmarks and compare it to state of the art storage policies when using a limited memory budget to recall. Using MOE, we find that we consistently surpass random selection and even more complex baselines, such as herding \cite{rebuffi2017icarl}. Thus, our contributions can be summarized as follows:
\begin{itemize}
    \item We introduce MOE, a storage policy that removes outliers and noisy samples from memory by verifying the label-homogeneous.
    \item Through a series of experiments, we demonstrate that MOE surpasses SOTA policies in scenarios with memory buffer constraints.
    \item We develop an understanding of why it is important to remove outliers by performing an ablation study.
\end{itemize}


\section{Background: Problem Formulation}

\subsection{Continual Learning}
In CL, we consider a stream of $T$ tasks. Each task $t$ consists of a new data distribution $D^t = (X^t,Y^t)$, where $X^t$ denotes the input instances and $Y^t$ denotes the instance labels. The goal is to train a classification model $ f_{\Theta}: X \longrightarrow Y $ using data from a sequence of $T$ tasks: $ D = \{D^1, ..., D^T \} $. During each task, model $f_{\Theta}$ will minimize the objective function $\mathcal{L}$ using data $D^t$.

\begin{equation}
    \mathcal{L}(D^t) = \frac{1}{N^t} \sum_{t=1}^{N^t} \mathcal{L}_t (f_{\Theta}(x_i^t), y_i^t)
    \label{eq:cl}
\end{equation}

Each task is presented sequentially to the model and trained for $E$ epochs. We focus on a Class-Incremental (Class-IL) setting, as has been the main focus of recent CL scientific endeavors. Such a scenario is much more challenging and realistic than the traditional Task Incremental (Task-IL) setting \cite{van2019three}. Unlike the Task-IL, a task descriptor is only available during training in Class-IL.

For the case of memory-based methods, together with minimizing Equation \ref{eq:cl}, model $f_{\Theta}$ needs to minimize $\mathcal{L}$ using the data available in memory $M$ at time $t$. The buffer $M^t$ comprises $|M|$ samples from previous distributions, meaning that at task $t$, the buffer will contain samples only from $t'<t$. 

\begin{equation}
    \mathcal{L}_M(D^t, M^t) = \mathcal{L}(D^t) + \frac{1}{|M|} \sum_{i=1}^{|M|} \mathcal{L}_t (f_{\Theta}(m_i^t, y_i^t) 
    \label{eq:cl_memory}
\end{equation}

Current memory-based methods have achieved promising results on many CL benchmarks. Saving examples helps mitigate forgetting from previous tasks by representing past distributions used during training. For the memory to be sufficient, it must represent the previous distribution as fully as possible, considering all its classes and concepts.
The function in charge of populating memory $M$ is known as \textit{Storage Policy} and decides which elements go into the memory by sampling from set $D^t$ given a function $\mathbf{P}$, as shown in Equation \ref{eq:storage_policy}. An ideal policy function is the one that minimizes Equation \ref{eq:cl_memory} for evaluation stream $D^1 ... D^T$, restricted by the memory size $|M|$. 

\begin{equation}
    M^{t+1} \leftarrow \mathbf{P}(M^t, D^t), \quad |M^{t+1}| \leq |M|
    \label{eq:storage_policy}
\end{equation}

In most cases, we assume that $M^{t+1}$ will always contain $|M|$ samples, and the storage policy will decide which samples to remove to add those from new task $D^t$.

\subsection{Datasets as Mixture of Subpopulations}
Previous studies in memorization of deep learning models have discussed the idea of data distributions being mixtures of subpopulations within the data \cite{Feldman20}. For instance, for a class of images with the label "bird," subpopulations might represent different bird species, camera angles, or lighting situations. This point of view will be fruitful for analyzing the different storage policies ($\mathbf{P}$) compared in this work, as all of them can be interpreted as different ways of \textit{choosing representatives for one or more subpopulations} in the data distribution.

There could be many ways to determine the subpopulations in the data. For simplicity, we approximate them as \textit{neighborhoods in embedding space}. By doing this, we can analyze two features of subpopulations: their size and label-homogeneity. These allow us to understand how representative a population is and whether it is worth training on. Size is desirable as it helps identify subpopulations that are more representative of the whole distribution, while label-homogeneity allows us to distinguish subpopulations that may be noisy or outliers. A neighborhood around a sample that contains many elements from different classes suggests that the sample would be close to a complex part of the decision boundary of the model \cite{Arplt2017}. This is undesirable for a storage policy as it will require much training to learn those boundaries correctly. One must note that while focusing on complex parts of a decision boundary might help a model reach high precision, in our setting, we are looking for the least amount of data points that can keep a consolidated memory of past tasks. Hence we look for candidates in the consolidated subpopulations of our data.

Ideally, memory $M$ should represent the majority of subpopulations that make up the distribution of the training data for each task. However, it is common to have limitations on the number of elements we can save. Therefore, it is necessary to select which samples to save and from which subpopulations to select. This becomes more important as some of the smaller subpopulations can represent outliers or mislabeled samples, which can generate noisy representations, affecting future training.

\section{Memory Outlier Elimination (MOE)}

\begin{figure*}
  \centering
  \includegraphics[width=0.95\linewidth]{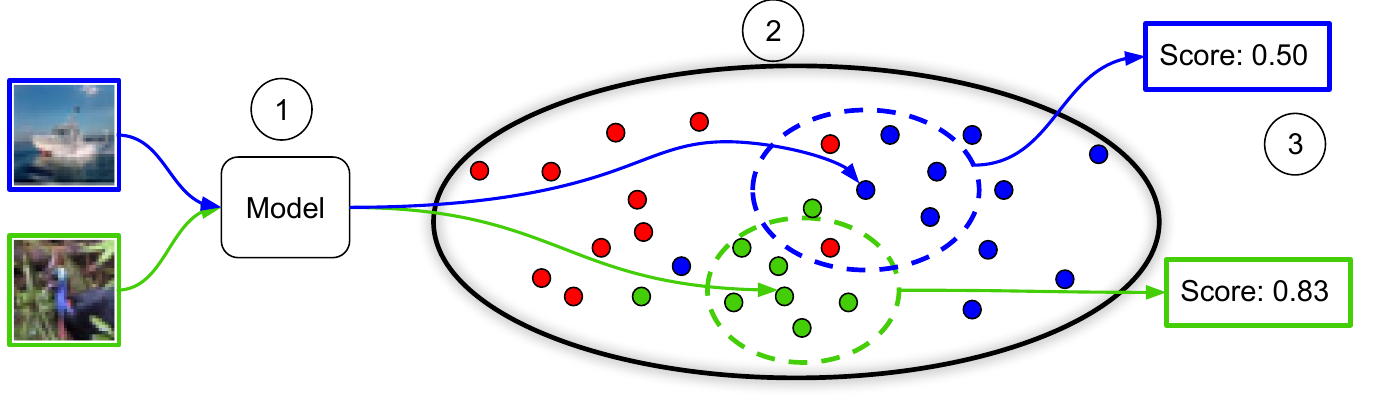}
    \caption{To calculate the label-homogeneity of a sample, first (1) we must obtain a feature vector for each sample of the task. These vectors come from a pre-trained model or the current model, which is a design decision. Then, for each sample, the $N$ nearest neighbors (2) - using a cosine distance - are retrieved to count how many of those belong to the same sample class (3).}
    \label{fig:explain_proxy}
\end{figure*}

We hypothesize that current functions $\mathbf{P}$ are hampered by selecting outliers and noisy samples, which limit their capacity for generalization when faced with memory constraints. Therefore, we propose a method to eliminate these outliers from the sampling pool.

To achieve this elimination, we propose to look at the labels of a given sample's neighbors in an embedding space. In particular, we calculate the ratio of neighbors that share the same label as the sample. For a given sample $\{x, y\}$ we calculate its label-homogeneity $H$ against its nearest $N$ neighbours $\{x_i, y_i\}, i=1..N$, following Equation \ref{eq:score}. A similar metric has been used in the Curriculum Learning literature as a proxy for Learning Consistency \cite{jiang2021characterizing}. A visual approximation can be found in Figure \ref{fig:explain_proxy}.

\begin{equation}
    \label{eq:score}
   H(x,y) = \frac{1}{N}\sum \limits_{i=1}^N 1[y = y_i].
 \end{equation}

We then proceed to eliminate from the memory sampling pool those samples that are below a threshold $H'$. Finally, MOE selects randomly from the remaining samples. In this way, MOE selects from a significant number of consolidated subpopulations of the data. We restrict the policy to maintain the same amount of samples per class as in most commonly used in previous storage policies functions $\mathbf{P}$.

\begin{equation}
    M^{t+1} \leftarrow \mathbf{P}(M^t, D^t | H(x,y) \geq H')
\end{equation}

To obtain the label-homogeneity, we generate the embedding space using the currently trained model. It is important to note that pre-trained models can also be used to obtain the embedding space. However, this assumes the need for a pre-trained model with characteristics similar to the actual data.

\begin{figure*}
    \centering
    \includegraphics[width=0.95\linewidth]{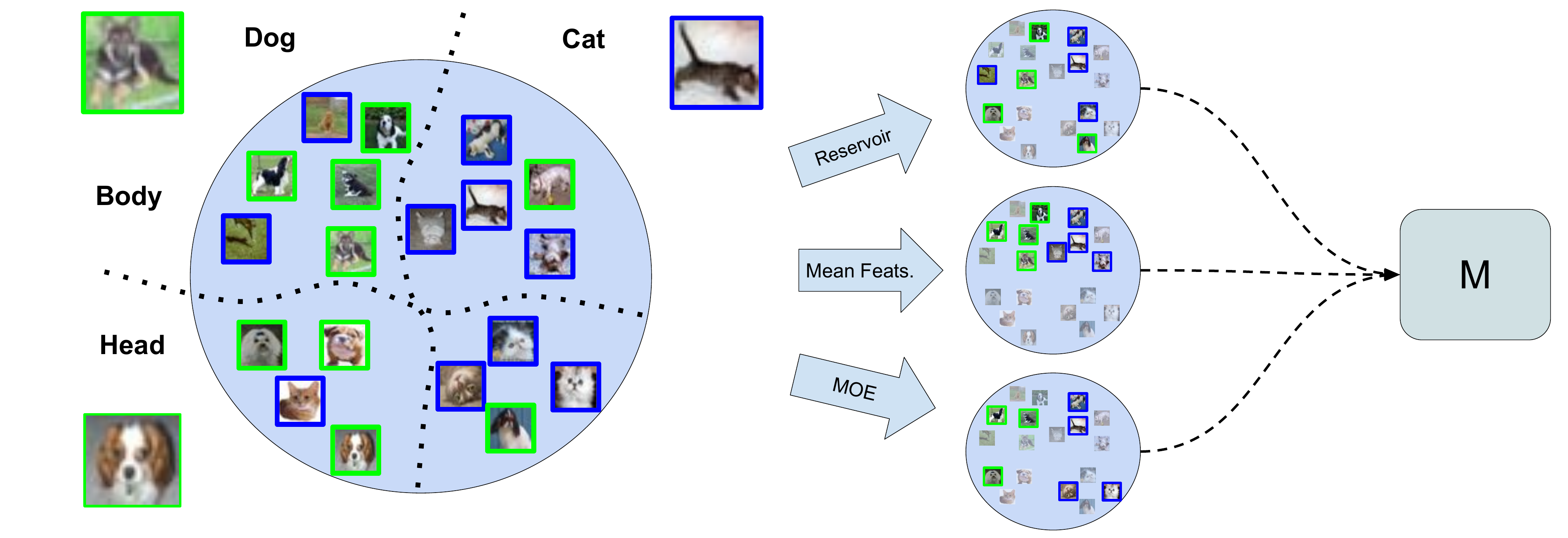}
    \vspace{1mm}
    \caption{Diagram of different memory population strategies. The Reservoir method considers representatives from all subpopulations, even if they are outliers. Mean of Features limits the number of subpopulations to the number of classes and samples from around class means which may limit its access to parts of the data distribution. MOE, on the other hand, samples from all subpopulations that are not outliers.}
    \label{fig:explain_moe}
\end{figure*}

By selecting examples with highest label-homogeneity, the memory will be mostly populated with examples from the most significant subpopulations. This helps to obtain easy-to-learn and easy-to-remember samples from a small group of subpopulations of a class, reducing diversity that can affect the representation of the class distribution. To add diversity, MOE randomly samples from these high label-homogeneity samples. A visual explanation can be found in Figure \ref{fig:explain_moe}.

\section{Experimental Setup}
To clearly reflect the contribution of our approach, we adopt a simple ER and test our proposal by training on 1, 5, or 10 epochs per task. This is done for two reasons: 1) Computational constraints: we are testing multiple memory population methods, datasets, and hyperparameters. 2) We tested training models for longer and found that in most cases the improvements were marginal, and the results we observed for fewer epochs also held for this configuration. 

\subsection{Datasets}
We train models on different sequences of CIFAR-10 and CIFAR-100 \cite{krizhevsky2009learning} split into 5 tasks, and CORe50 \cite{pmlr-v78-lomonaco17a} in 9 tasks (following the splits outlined in \cite{maltoni2019continuous}). CIFAR-10 and CIFAR-100 datasets are traditionally used in CL, while we also train on CORe50 to understand how methods behave with more complex datasets.

\subsection{Baselines}
Many approaches have been proposed for populating memory in CL, but few studies have compared the effectiveness of these approaches under different conditions. Some research has even shown that, in certain circumstances, there is little difference between different methods, highlighting how limited our understanding of replay strategies is. In this work, we will use the following methods as baselines:

\paragraph{(a) Reservoir.} The Reservoir method \cite{10.1145/3147.3165} allows elements to be selected from a stream without knowing how many instances to expect. It works by selecting each sample with a probability $\frac{M}{N}$, where $N$ is the number of elements observed so far and $M$ is the size of the memory. This approach randomly maintains a uniform sample from the observed stream. The Reservoir strategy selects representative samples from different subpopulations without discriminating over their size or noise.

\paragraph{(b) Class Balance (CBRS).} This method is similar to the Reservoir strategy but ensures that each class is represented equally in the memory buffer \cite{chrysakis2020online}. We use a dynamic assignment, meaning that the memory is always full. Samples of new classes replace instances of old classes to maintain equal representation in the memory.

\paragraph{(c) Min/Max Loss.} This method selects samples associated with lesser or greater loss values. It is a proxy for easier or harder samples. This is a storage policy reminiscent of what would be done for Self Paced Curriculum \cite{spcl} or Anticurriculum, respectively.

\paragraph{(d) GSS.} This method \cite{aljundi2019gradient} selects samples that maximize the differences in gradient directions with other samples stored in the memory. It can be seen as a method that selects representatives from different subpopulations in \textit{gradient space}.

\paragraph{(e) Mean of Features (MF).} Also know as herding, this method, proposed by \cite{rebuffi2017icarl}, calculates an average class feature vector based on the representations of the elements in memory for a given class. If the distance between the new vector and the corresponding class vector is smaller than the farthest vector in the memory, the new example is replaced with the farthest one. This method considers a number of subpopulations equal to the number of classes for composing the memory centered on each class mean. This might also limit noisy samples or outliers, as these should be far from the class mean. However, by forcing a fixed number of neighborhoods, this method fails to find multiples relevant subpopulations. 

As standard practice in the replay-based method, the memory buffer is completely filled in the first task with data from $D^1$ to later update the memory with new data at the end of each task. The memory always keeps $|M|$ elements. Except for the Reservoir, all baselines have class-balanced constrain in the memory.
      
\subsection{Implementation Details}
All experiments are run with 3 different seeds, each inducing a different ordering of sequences. In the case of CIFAR-10 and CIFAR-100, we use a simple convolutional architecture proposed in \cite{mirzadeh2022architecture}. For CORe50, the simple convolutional architecture is insufficient, and we resort to using a reduced version of Resnet-18 architecture proposed in \cite{rebuffi2017icarl}, which is standard for these datasets in CL. 
The optimizer is SGD with a learning rate of $0.001$, momentum $0.9$, and batch size of $32$ unless otherwise mentioned. All methods are implemented and trained using Avalanche \cite{lomonaco2021avalanche}. Using Avalanche, we have also developed extra functions available in \url{github.com/JuliousHurtado/MOE}.

We evaluate the average accuracy over the $T$ tasks after the sequential learning (Acc). We also measure forgetting (For), which tells us how much performance is lost on previous tasks after sequential learning. These metrics were both proposed by \cite{lopez2017gradient}. Equation \ref{eq:metrics_acc} and \ref{eq:metrics_for} shows the formulas for the metrics, where $A_{i,j}$ is the accuracy of task $i$ after training task $j$. 

\begin{equation}
Acc = \frac{1}{T} \sum_{i=1}^T Acc_{T,i}
\label{eq:metrics_acc}
\end{equation}

\begin{equation}
For = \frac{1}{T-1} \sum_{i=1}^{T-1} Acc_{T,i} - Acc_{i,i}
\label{eq:metrics_for}
\end{equation}

\section{Results}

\begin{table*}[] 
\begin{center}
\small
\renewcommand{\arraystretch}{1.15}%
\setlength{\tabcolsep}{3pt}
\begin{tabular}{l|ccc|ccc|ccc}
  & \multicolumn{3}{c}{CIFAR-10} & \multicolumn{3}{c}{CIFAR-100}& \multicolumn{3}{c}{CORe50}\\
\# Epochs     & 1 & 5 & 10 & 1 & 5 & 10 & 1 & 5 & 10 \\
\hline
Reservoir & 26.3±0.6\% & 22.1±1.7\% & 22.9±1.1\% & 11.0±0\% & 14.9±0.3\% & 15.6±0.3\% & 21.6±1.1\% & 25.0±1.0\% & 24.8±0.8\% \\
GSS       & 21.6±1.3\% & 20.0±1.9\% & 20.1±1.8\% & 8.9±1.1\%  &12.1±0.6\%& 13.6±0.4\% & 15.6±0.8\% & 12.0±1.2\% & 12.4±1.7\% \\
CBRS      & 25.8±0.7\% & 22.2±1.5\% & 23.1±1.3\% & 11.5±0.3\% &14.9±0.5\%& 15.9±0.3\% & 23.3±0.7\% & 24.3±0.7\% & 25.0±1.1\% \\
Max Loss  & 17.1±0.2\% & 18.4±0.4\% & 18.8±0.2\% & 5.8±1.2\%  & 10.4±1.5\%& 11.1±1.4\%& 11.3±1.9\% & 12.5±2.0\% & 15.2±0.9\% \\
Min Loss  & 20.7±1.0\% & 20.2±0.4\% & 20.7±0.5\% & 12.7±0.3\% & 16.3±0.6\%& 17.7±0.8\%& 17.7±2.5\% & 19.0±0.4\% & 22.1±2.1\% \\
MF        & 29.4±2.3\% & 25.2±2.0\% & \textbf{24.7±1.4\%} & 12.2±0.2\% &16.4±0.2\%& 17.7±0.9\% & \textbf{24.8±2.6\%} & \textbf{26.0±1.9\%}  & 26.1±0.5\% \\
\hline
MOE & \textbf{29.6±0.3\%} & \textbf{25.4±1.4\%} & \textbf{24.7±1.1\%} & \textbf{14.5±0.3\%} & \textbf{18.6±0.4\%} & \textbf{19.1±0.5\%} & 23.6±0.8\% & 25.3±0.6\% & \textbf{26.3±0.9\%}  
\end{tabular}
\vspace{3mm}
\caption{Accuracy for models trained with different storage policies for 1, 5, and 10 training epochs. Memories are populated with 5 elements per class. As can be seen, MOE equals or outperforms all baselines for all datasets. Best results in bold.}
\label{tab:tab_sec1}
\end{center}
\end{table*}

Table \ref{tab:tab_sec1} shows the mean accuracy across a sequence of tasks obtained by the baseline methods. Our results indicate that Mean of Features (MF) outperforms all other baseline methods, including the Reservoir strategy. This is relevant, as the Reservoir strategy is the default memory population method used in Memory-based methods, highlighting the importance of memory population methods for better performance. On the other hand, when applying MOE, we can see a clear increase in the accuracy achieved in all experiments. By removing the outliers of each class and then randomly sampling over those that remain, we can help a replay-based strategy to improve its results by only sampling from a pool that correctly represent the class.

\begin{table*}[t]
\begin{center}
\small
\renewcommand{\arraystretch}{1.15}%
\setlength{\tabcolsep}{3pt}
\begin{tabular}{l|ccc|ccc|ccc}
  & \multicolumn{3}{c}{CIFAR-10} & \multicolumn{3}{c}{CIFAR-100}& \multicolumn{3}{c}{CORe50}\\
\# Epochs     & 1 & 5 & 10 & 1 & 5 & 10 & 1 & 5 & 10 \\
\hline
Reservoir & 57.9±3.4\% & 78.1±5.1\% & 82.0±3.8\% & 14.1±0.2\% & 31.1±0.2\% & 41.1±0.3\% & -11.9±1.7\% & -13.8±1.1\% & -13.5±1.4\% \\
GSS       & 62.6±4.7\% & 83.0±4.8\% & 86.2±4.2\% & 15.8±0.4\% & 39.1±0.8\% & 47.1±0.9\% & -7.1±0.5\% & -2.2±0.3\% & -1.9±0.1\% \\  
CBRS      & 56.6±3.8\% & 77.4±4.1\% & 81.4±3.5\% & 13.9±0.4\% & 31.1±0.2\% & 40.6±0.5\% & -13.8±1.1\% & -13.3±1.3\% & -13.3±0.6\% \\   
Max Loss  & 69.8±5.4\% & 80.8±2.7\% & 84.8±3.7\% & 19.5±2.1\%  & 37.5±2.2\% & 46.6±1.9\% & -3.9±3.5\% & -3.9±2.3\% & -6.4±1.3\% \\
Min Loss  & 65.5±3.7\% & 80.5±2.5\% & 83.8±3.4\% & \textbf{7.0±1.9\%} & 29.8±1.4\% & 40.1±1.6\% & -10.0±2.6\% & -11.2±1.2\% & -12.5±1.0\% \\
MF        & 54.3±6.0\% & 74.3±5.6\% & \textbf{79.0±3.8\%} & 14.0±0.1\% & \textbf{29.2±1.6\%} & \textbf{38.1±1.3\%} & -11.9±2.6\% & -13.8±1.1\% & -13.5±1.4\% \\
\hline
MOE       & \textbf{52.8±1.6\%} & \textbf{73.6±5.2\%} & 79.6±3.8\% & 10.7±1.5\% & 31.1±1.1\% & 40.7±1.6\% & \textbf{-14.7±0.7\%} & \textbf{-14.4±1.3\%}  & \textbf{-14.3±1.3\%}   \\  
\end{tabular}
\vspace{3mm}
\caption{Forgetting for models trained with different storage policies for 1, 5, and 10 training epochs. Memories are populated with 5 elements per class. MOE outperforms or is on par with competing methods, suggesting that performance gains from MOE come from both rapid relearning and less forgetting. Best results in bold, second best underlined.}
\label{tab:tab_sec1_for}
\end{center}
\end{table*}

Additionally, we observed that MOE works better as we increase the number of training epochs. By increasing the gap against the baselines, it demonstrate the power of generalization of the selected samples and not only of adaptation. However, we observed that on CIFAR-10, we obtain worse results as we train for more epochs. This can be explained because CIFAR-10 is a relatively simple dataset, and forgetting occurs very rapidly when using a small memory size, as can be seen Table \ref{tab:mem_size_abl}.

We see that MF outperforms or is competitive with MOE only on CORe50. But we argue that this dataset plays to the strengths of MF. CORe50 consists of video frames from objects in different settings or sessions. Thus, image samples from this dataset are pretty similar to one another, therefore their embeddings will be close, which is ideal for a centroid-based method like MF. Even in this setting, MOE is competitive or outperforms MF when we increase the number of epochs. 

It is critical to note the robustness of MOE compared to other methods, especially MF. This can be seen by comparing the standard deviation obtained by the methods. MOE has much lower values, showing its stability. This shows that outlier removal helps performance but also can help mitigate noise during training.

Similar to accuracy, of all the baselines the one that forgets the least is MF, as can be observed in Table \ref{tab:tab_sec1_for}. The fact that selecting the right samples to be stored in memory enhances learning and reduces forgetting. As in the previous results, here we can also see how MOE outperforms other methods in most cases. The negative value of CORe50 indicates that the model continues to learn from previous tasks.

\subsection{Increasing Memory Size}
We also test MOE on settings with increased memory sizes while still working in a constrained memory size environment ($\leq 20\%$ dataset size for CIFAR-100, $\leq 0.4\%$ for CIFAR-10, $\leq 0.8\%$ for CORe50). Table \ref{tab:mem_size_abl} shows these results. We observe that for 10 and 20 samples per class, MOE still outperforms MF consistently, but as we increase memory size, the gap between MOE and MF decreases. This suggests that MOE is more capable of using memory more efficiently than MF when severely constrained in memory, but as we increase the size of the memory, this advantage decreases as both methods converge to a similar sampling space. This can also explain previous study that show that there is no significant difference between reservoir and more complex methods when the memory is big enough to fully represent previous distributions \cite{wu2019large, hayes2020remind,araujo-etal-2022-mem}.

\begin{table*}
    \small
    \centering
    \renewcommand{\arraystretch}{1.15}%
    \setlength{\tabcolsep}{3pt}
    \begin{tabular}{l|ccc|ccc|ccc}
    & \multicolumn{3}{c}{CIFAR-10} & \multicolumn{3}{c}{CIFAR-100} & \multicolumn{3}{c}{CORe50} \\
    Samples per Cls. & 5 & 10 & 20 & 5 & 10 & 20 & 5 & 10 & 20\\ 
    \hline
    Reservoir   & 22.9±1.1\% & 26.9±1.1\% & 31.0±0.9\% & 15.6±0.3\% & 19.3±0.6\% & 23.4±0.7\% & 24.8±0.8\% & 27.3±0.8\% & 31.6±0.9\%\\
    MF          & \textbf{24.7±1.4\%} & 29.4±2.5\% & 34.4±0.9\% & 17.7±0.9\% & 20.6±0.4\% & \textbf{26.8±0.7\%} & 26.1±0.5\% & 29.5±1.1\% & 33.0±1.2\%\\
    MOE   & \textbf{24.7±1.1\%} & \textbf{30.3±3.7\% }& \textbf{37.7±0.8\%} & \textbf{19.1±0.5\%} & \textbf{21.9±0.7\%} & 25.6±0.4\% & \textbf{26.3±0.9\%} & \textbf{30.0±1.2\%} & \textbf{33.4±0.8\%} 
    \end{tabular}
    \vspace{3mm}
    \caption{Accuracy for different memory population methods when increasing memory size for 10 training epochs. Results for 5, 10, and 20 samples per class. We observe that MOE still outperforms MF when increasing memory size but the performance gap is reduced as memory grows.}
    \label{tab:mem_size_abl}
\end{table*}

\subsection{Limitations and Mitigations}

\begin{figure*}
\begin{tabular}{cc}
\includegraphics[width=0.48\linewidth]{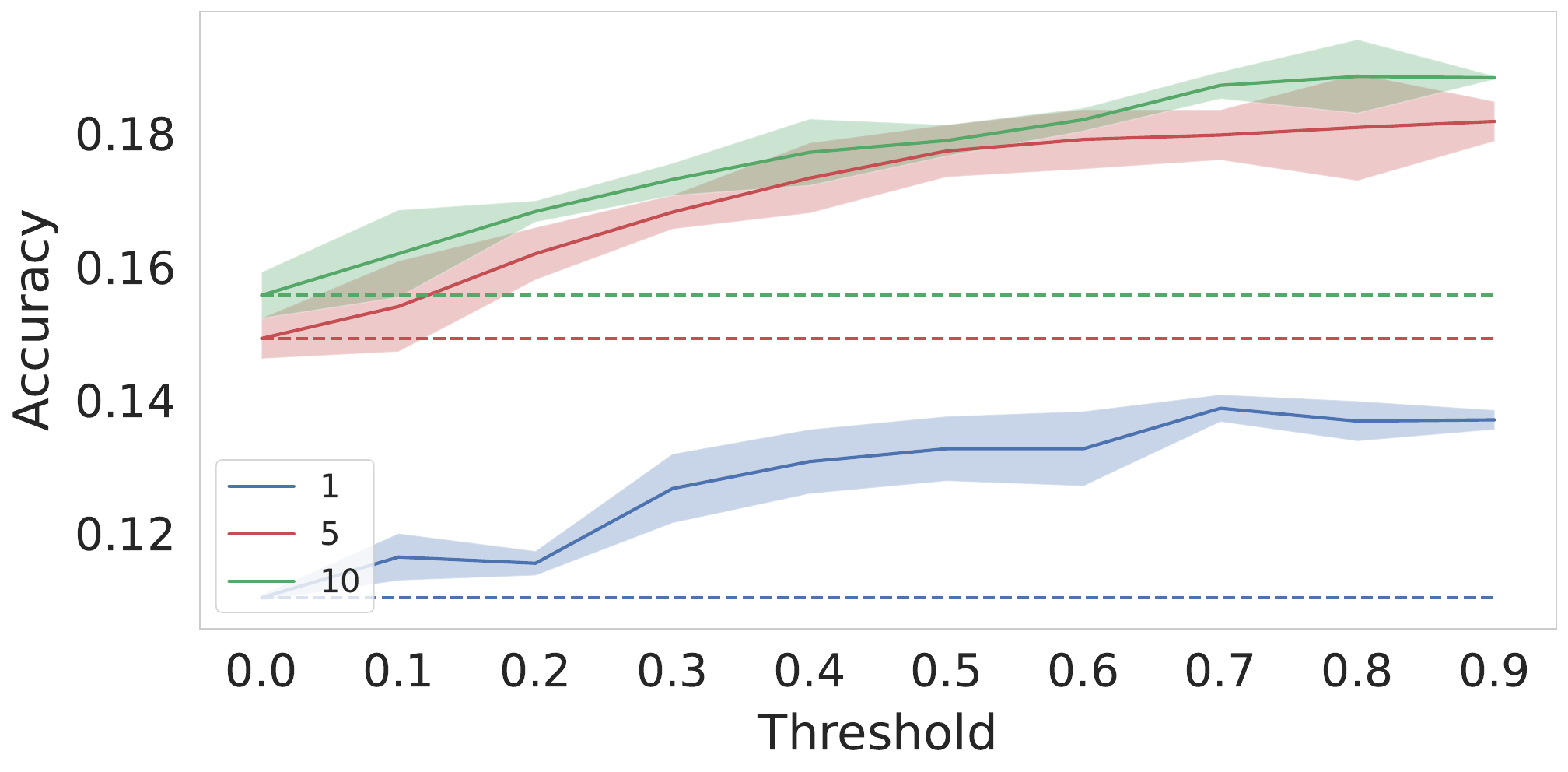} \label{fig:abl_cifar100_thres}&
\includegraphics[width=0.48\linewidth]{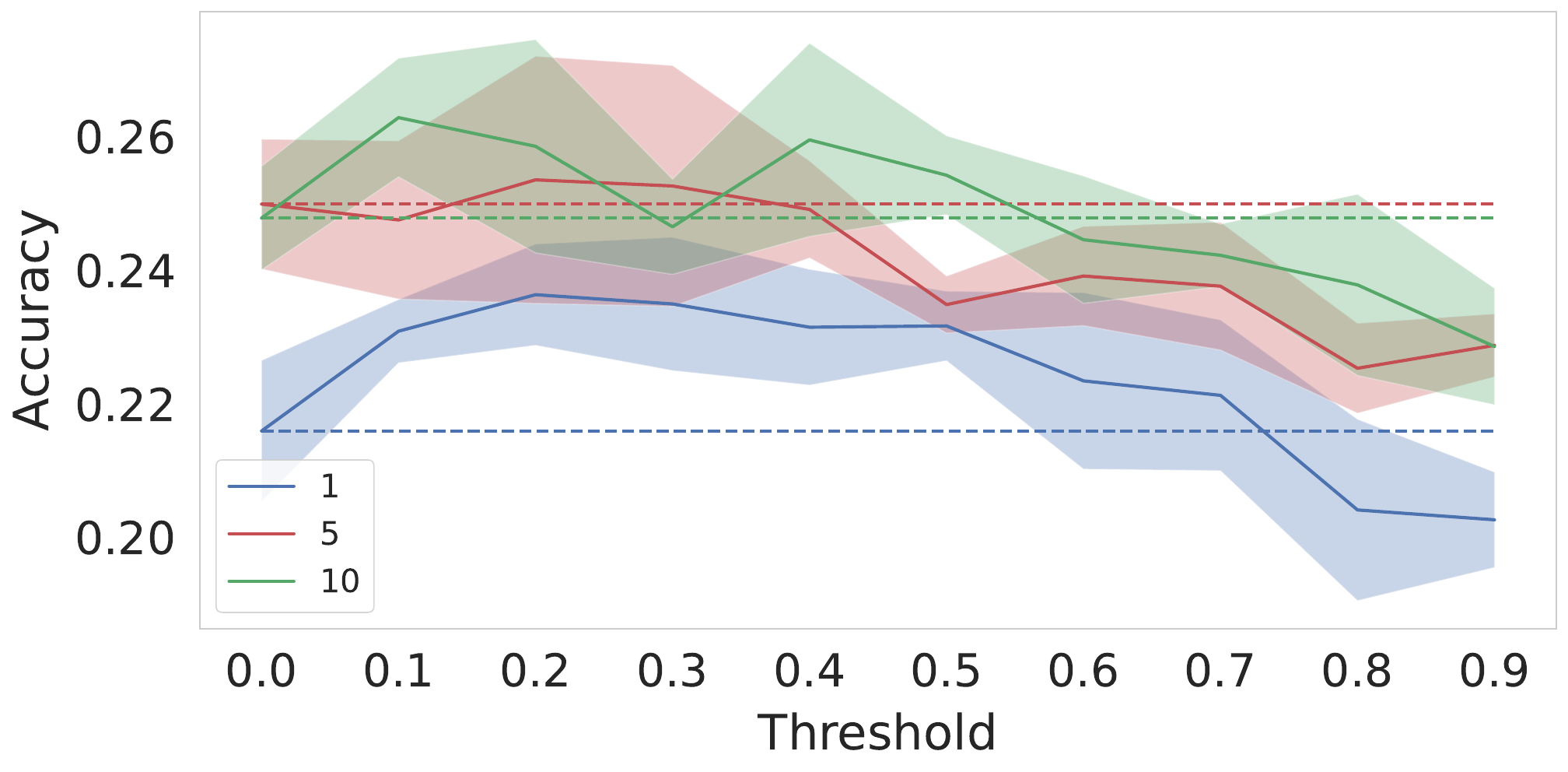} \label{fig:abl_core50_thres}\\
CIFAR-100&CORe50 \vspace{2mm}
\end{tabular}
\caption{Accuracy achieved by MOE for different thresholds when training for different amounts of epochs on CIFAR-100 and CORe50. The dotted lines represent the Reservoir method. While displaced in the y-axis, accuracy curves show relatively similar shapes, which shows evidence that an optimal threshold for one level of computation should work well with other levels of computation. This suggests a strategy for finding the optimal threshold by training for 1 epoch on different thresholds.}
\label{fig:abl_thres}
\end{figure*}

A limitation of our proposed method is the requirement for determining an appropriate label-homogeneity cutoff point. Through experimental analysis, we investigated the performance of models trained to utilize MOE for varying numbers of training epochs. Figure \ref{fig:abl_thres} shows accuracy results for different thresholds when training models for varying amounts of epochs. Our results indicate that the accuracy as a function of threshold exhibits similar characteristics. In particular, the curves are similar in shape but displaced in the accuracy axis. Therefore, to mitigate this issue, we propose a strategy of selecting the cutoff point through training for a single epoch with various threshold values. This approach is analogous to the common practice of selecting hyperparameters, such as the learning rate and network architecture, in neural network training.

\section{Ablations}

It is important to note that MOE assumes that the representation space is distributed in a mixture of sub-populations, meaning that each class can be represented by multiple sub-distributions. To test this hypothesis, we propose different versions of MOE, which enforce certain ablations of our method either on the embedding space or the sampling method:
\begin{itemize}
    \item \textbf{MOE-Upper}: we deterministically select the \textit{most homogeneous} samples per class.
    \item \textbf{MOE-Lower}: we deterministically select the \textit{least homogeneous} samples per class.
    \item \textbf{MOE-ImageNet:} we embed samples using a Resnet-18 pretrained on ImageNet, then we sample randomly from these. This version is meant to test how relevant a good embedding is for MOE.
    \item \textbf{MOE-Model}: the vanilla MOE. We embed samples using the currently trained model, then we sample randomly from these. This setting is more realistic as there may not be a pretrained model available for a given task.
\end{itemize}

\subsection{Label-homogeneity}

The first characteristic to verify is that making a random selection over the group of elements over a threshold $H'$ is better than always selecting the ones with the greatest or lower homogeneity (MOE-Upper and MOE-Lower). If there are indeed several sub-populations for each class, it is necessary to diversify where we sample the data from, and by selecting only the most homogeneous, it is most likely that we will sample very similar data.

\begin{figure*}
\begin{tabular}{cc}
\includegraphics[width=0.48\linewidth]{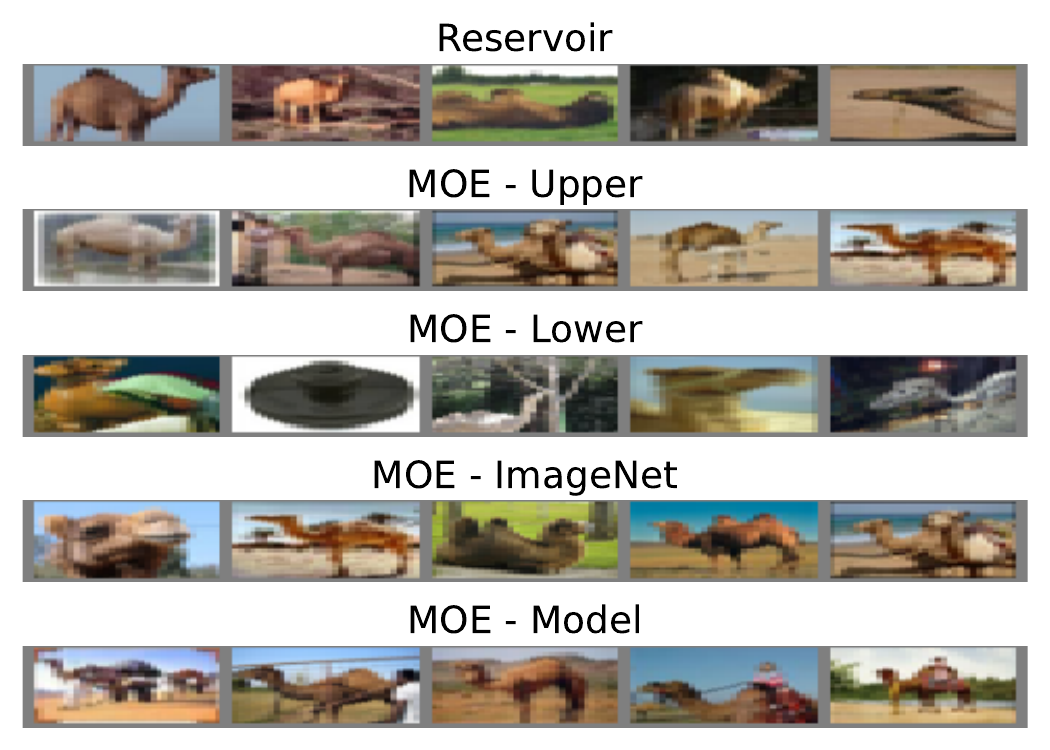} \label{fig:mem_comp_cls_15}&
\includegraphics[width=0.48\linewidth]{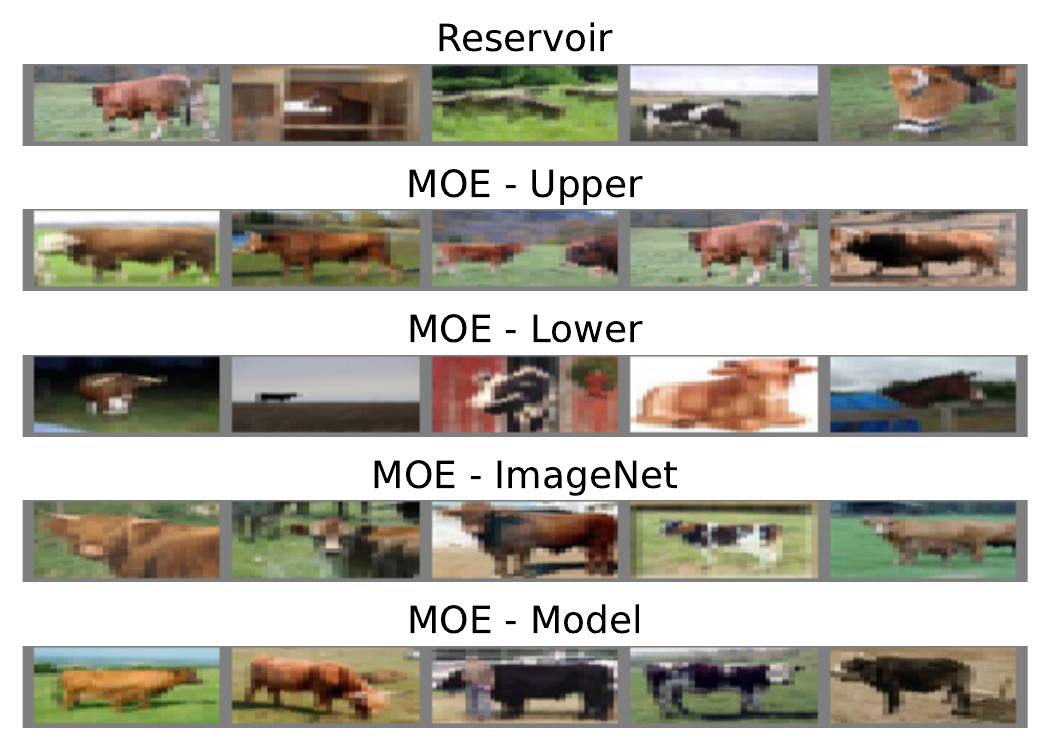}\label{fig:mem_comp_cls_19}\\
Memory CIFAR-100 - Class Camel&Memory CIFAR-100 - Class Cow \vspace{2mm}
\end{tabular}
\caption{Memory Samples from CIFAR-100 for different memory population methods. We observe that randomly choosing samples produces more varied samples including outliers. MOE-Upper produces mostly canonical versions of both camels and cows. MOE-Lower presents diverse but noisy or outlier samples. MOE-ImageNet and MOE-Model produce diverse samples like the Reservoir method but without any of the outliers.}
\label{fig:mem_comp}
\end{figure*}

As expected, we found that MOE-Lower performed poorly, as it tends to choose outliers and noisier samples which do not aid in generalization. This can be clearly seen in Figure \ref{fig:mem_comp}, where MOE-Lower chooses noisy or outlier samples, where camera angles and colors are outside the norm for the class. However, the other three versions of MOE tend to outperform MF or be competitive with it, as shown in Table \ref{tab:ablation_moes}. On the other hand, MOE-Upper remarkably outperforms previous baselines, however, lags behind MOE. The lack of diversity in the selection of MOE-Upper is clearly shown in Figure \ref{fig:mem_comp}, as most images selected represent a very similar concept, which can help in simple benchmark (CIFAR10), but it is strongly affected in more complex datasets (CORe50).

\begin{table*}[ht]
\begin{center}
\small
\renewcommand{\arraystretch}{1.15}%
\setlength{\tabcolsep}{3pt}
\begin{tabular}{l|ccc|ccc|ccc}
  & \multicolumn{3}{c}{CIFAR-10} & \multicolumn{3}{c}{CIFAR-100}& \multicolumn{3}{c}{CORe50}\\
\# Epochs     & 1 & 5 & 10 & 1 & 5 & 10 & 1 & 5 & 10 \\
\hline
MOE-Upper & 28.3±1.1\% & 25.0±1.3\% & 25.3±0.9\% & 13.5±0.5\% & 18.0±0.2\% & 19.0±0.2\% & 22.1±0.4\% & 22.0±0.5\% & 21.5±0.5\% \\
MOE-Lower & 19.9±0.4\% & 19.6±0.3\% & 20.0±0.6\% & 9.1±0.2\% & 11.9±0.3\% & 13.0±0.1\% & 22.3±0.6\% & 23.8±0.7\% & 23.4±0.4\% \\
MOE-ImgNet & 28.6±2.3\% & 26.0±1.4\% & 25.2±1.8\% & 15.0±0.2\% & 18.6±0.3\% & 19.2±0.6\% & 24.7±0.7\% & 26.0±0.9\% & 26.9±0.6\% \\
MOE-Model & 29.6±0.9\% & 25.4±1.4\% & 24.7±1.1\% & 14.5±0.3\% & 18.6±0.4\% & 19.1±0.5\% & 23.6±0.8\% & 25.3±0.6\% & 26.3±0.9\%  
\end{tabular}
\vspace{3mm}
\caption{Performance of different ablations of MOE. Choosing the least homogeneous samples incurs heavy penalties on performance, while chossing only the most homogeneous works better but still lags behind MOE. While using a pretrained model to extract features is better as exoected, MOE is still competitive.}
\label{tab:ablation_moes}
\end{center}
\end{table*}

\subsection{Embedding quality}

The second characteristic to check is how good are the current model's embeddings for calculating label-homogeneity, as we cannot guarantee having a pretrained model for our specific tasks. To test this, we compare MOE against a version that uses a Resnet18 pre-trained on ImageNet to obtain the features vectors. We can see that despite the advantage of having a pre-trained model, MOE-ImgNet does not significantly outperform MOE. We can even see in Table \ref{tab:ablation_moes} that MOE's accuracy is higher or equal in some cases.

\section{Related work}
Memory-based methods address catastrophic forgetting by incorporating data from previous tasks into the training process for the current task \cite{ebrahimi2021remembering, buzzega2021rethinking}. These approaches can use raw samples \cite{rebuffi2017icarl, chaudhry2019tiny}, minimize gradient interference  \cite{lopez2017gradient, chaudhry2018efficient}, or train generative models such as GANs or autoencoders \cite{lesort2019generative, shin2017continual, kemker2018fearnet} to generate samples from previously seen distributions.

Multiple approaches for populating the memory in memory-based CL methods exist. One simple but effective method is the Reservoir strategy \cite{10.1145/3147.3165}, which selects elements at random. Other strategies have been proposed that use various metrics to choose more representative elements for the memory \cite{chaudhry2019tiny, hayes2020remind, hayes2021selective, aljundi2019gradient}. Some research has focused on the impact of hyperparameters on certain methods \cite{merlin2022practical} or the effect of rehearsal methods on loss functions  \cite{verwimp2021rehearsal}. Other studies have explored methods for selecting elements from the memory, such as selecting elements based on how much their loss would be affected  \cite{aljundi2019online} or using a ranking based on the importance of preserving prior knowledge \cite{isele2018selective}.

Despite the widespread use of memory-based methods in CL, the impact of memory composition on these methods has received relatively little attention \cite{tiwari2022gcr}. Some approaches in this area have employed Reservoir strategies \cite{chrysakis2020online} or used entropy-based functions to increase memory diversity \cite{wiewel2021entropy, sun2021information}. Others have focused on minimizing the angles between gradients for different elements in the memory to increase diversity \cite{aljundi2019gradient}. While these approaches have shown promise in certain situations, few studies have specifically targeted improving the representativeness of the memory.

A related field is that of Coreset construction, which entails finding a subset of a dataset that can achieve performance similar to using the whole dataset. Some works\cite{NEURIPS2020_aa2a7737} pose the problem as a bi-level optimization but constrained to small memory sizes. Others choose these sets by finding elements that approximate gradients from the whole dataset\cite{tiwari2022gcr}.

\section{Conclusions and Future Work}

This work propose and examines the impact of a new memory storage policy, Memory Outlier Elimination (MOE), on Experience Replay (ER). MOE identifies representative samples to improve performance by finding a storage policy that accurately represents the most relevant subpopulations within the data distribution. We tested MOE against several state-of-the-art baselines and found it consistently outperformed or performed comparably. Our analysis showed that selecting samples in a label-homogeneous neighborhood in embedding space improved performance, especially when combined with the Reservoir strategy. Qualitatively, we observed that MOE effectively balances sample diversity while minimizing the presence of outliers and noisy samples. Furthermore, when testing MOE - which samples from a variable number of subpopulations - against a state-of-the-art method - which samples from a constant number of subpopulations - we found that the performance gap between the two methods widened as we increased memory size. This suggests positive evidence that the subpopulation point of view of memory construction is valid. Additionally, we provide guidance on selecting appropriate hyperparameters for MOE. Future work will focus on determining optimal label-homogeneity thresholds for specific tasks and datasets.

\paragraph{Acknowledgements:} We acknowledge the support from the Chilean Government funding through ANID and National Doctoral Scholarships. Also, this research has been financially supported by the National Center for Artificial Intelligence CENIA FB210017, Basal ANID.

{\small
\bibliographystyle{ieee_fullname}
\bibliography{egbib}
}

\end{document}